\documentclass[letterpaper, 10 pt, conference]{ieeeconf}  

\IEEEoverridecommandlockouts                              

\usepackage{cite}
\usepackage{amsmath,amssymb,amsfonts}
\usepackage{algorithmic}
\usepackage{graphicx}
\usepackage{textcomp}
\usepackage{xcolor}

\usepackage{siunitx}
\usepackage{multirow}
\usepackage{booktabs}

\title{\LARGE \bf
A Deeply Supervised Semantic Segmentation Method Based on GAN}

\author{Wei Zhao$^{1\dagger}$, Qiyu Wei$^{1\dagger}$, Zeng Zeng$^{*}$%
\thanks{$^{1}$Wei Zhao, Qiyu Wei, Zeng Zeng are with School of Microelectronics, Shanghai University (SHU), China, 201800 (Email: {zw1315753898, qywei}@shu.edu.cn, zeng\_zeng@hotmail.com)}%
\thanks{$^{\dagger}$Contributed equally to this work}%
\thanks{$^{*}$Corresponding auther}%
}

\begin{document}

\maketitle
\thispagestyle{empty}
\pagestyle{empty}

\begin{abstract}

In recent years, the field of intelligent transportation has witnessed rapid advancements, driven by the increasing demand for automation and efficiency in transportation systems. Traffic safety, one of the tasks integral to intelligent transport systems, requires accurately identifying and locating various road elements, such as road cracks, lanes, and traffic signs. Semantic segmentation plays a pivotal role in achieving this task, as it enables the partition of images into meaningful regions with accurate boundaries.
In this study, we propose an improved semantic segmentation model that combines the strengths of adversarial learning with state-of-the-art semantic segmentation techniques. The proposed model integrates a generative adversarial network (GAN) framework into the traditional semantic segmentation model, enhancing the model's performance in capturing complex and subtle features in transportation images.
The effectiveness of our approach is demonstrated by a significant boost in performance on the road crack dataset compared to the existing methods, \textit{i.e.,} SEGAN. 
This improvement can be attributed to the synergistic effect of adversarial learning and semantic segmentation, which leads to a more refined and accurate representation of road structures and conditions. The enhanced model not only contributes to better detection of road cracks but also to a wide range of applications in intelligent transportation, such as traffic sign recognition, vehicle detection, and lane segmentation.

\end{abstract}

\section{Introduction}
Deep learning has demonstrated enormous potential for improving many aspects of intelligent transportation applications, including traffic flow prediction, object recognition, traffic safety, route mapping, and vehicle localization. Intelligent Transportation Systems (ITS)~\cite{figueiredo2001towards,papadimitratos2009vehicular,cc1} are transforming navigation experiences in cities and on highways by integrating cutting-edge technologies, such as machine learning, computer vision. Image processing and analysis are critical components of ITS, enabling the detection, recognition, and classification of vehicles, pedestrians, and road damage. Semantic segmentation is a popular technique in computer vision and has drawn considerable attention in ITS field due to its proficiency in real-time segmentation and classification. Overall, through the integration of emerging technologies such as deep learning, ITS has the potential to significantly improve our daily transportation experiences, making them safer, more efficient, and more reliable.

The incorporation of semantic segmentation in ITS~\cite{oztel2021vision,wan2020intelligent,kaushik2020leveraging} is a rapidly progressing domain, with continuous research and development initiatives that striving to enhance the precision, resilience, and efficiency of these algorithms. Semantic segmentation, by leveraging machine learning and computer vision to divide images into sections based on semantic meaning of objects contained in the image, has the potential to revolutionize the way we interact with, operate, and design intelligent transportation systems. In intelligent transportation systems, the main goal of semantic segmentation is to equip machines with the ability to understand and interpret their visual surroundings. By accurately segmenting objects and their boundaries, it is feasible to identify and track various road elements, such as lanes, lane splits, signs, traffic lights, pedestrians, and other vehicles.

Road cracks are one of the road defects, and may cause vehicle instability, increased noise, and other problems\cite{othman2019road}. If not detected and repaired in time, the cracks may anlarge and worsen, potentially endangering traffic safety. Therefore, it is crucial to identify road cracks promptly and accurately. Research continues to be done to effectively solve this problem. Traditional computer vision methods have achieved some success in road crack recognition, but there are many limitations, such as the need to manually select and extract features and to set appropriate thresholds for judgment. This causes the algorithm to perform poorly in complex environments and leads it to require greater experience and skill from operators. In addition, traditional algorithms cannot adaptively learn the morphology and position of road cracks, and their generalization ability is weak.

In recent years, deep learning methods have demonstrated astonishing performance in various fields, and more and more researchers are focusing on how to use deep learning to solve road defect detection problems~\cite{singh2021highway,arena2020review}. Nowadays, a large number of deep learning methods have emerged in the field of road crack recognition, mainly image-based methods and point-cloud-based methods. Image-based methods typically use convolutional neural networks (CNNs) to automatically learn features, assist downstream segmentation, detect objects, and perform other tasks. Segmentation-based methods can accurately describe the severity and shape of cracks, while object-detection-based methods cannot. Point-cloud-based methods mainly use devices such as laser scanners or 3D cameras to obtain point cloud data of the road and use deep learning models for classification, segmentation, or regression. For example, models like PointNet~\cite{qi2017pointnet} and PointCNN~\cite{li2018pointcnn} can be used for road classification or segmentation. These methods~\cite{cc2,kashyap2022traffic} can directly process 3D data, avoiding some limitations in image processing but requiring greater computing resources and more complex data preprocessing.

In summary, segmentation-based methods are more suitable for road crack recognition tasks; they provide richer information for road supervisors, thus achieving more orderly road crack repair and ensuring better traffic safety. This work proposes a new framework that combines deep supervision and adversarial learning ideas with existing segmentation methods. The framework uses a deep supervision mechanism to implement hierarchical supervision of defect segmentation, with clear objectives at each level; this accelerates model convergence speed and enhances model interpretability. At the same time, we use adversarial learning to directly sample distributions, providing us the advantage of a theoretical approximation of actual data. We use the difference between the defect segmentation results and the actual target data distribution to further improve the effect of the deep supervision mechanism, thereby increasing the accuracy and robustness of the framework's defect recognition.

To sum up, the contribution of this work lies in three folds:
\begin{itemize}
\item We propose a simple and efficient structure that combines codec structure framework and adversarial learning, which trains the generator of the Unet network by means of adversarial learning. The structure is a general one that can be flexibly integrated into existing semantic segmentation neural networks.
\item We propose a way to use adversarial learning to help train semantic segmentation models and to address the shortcomings of previous semantic segmentation models applied in ITS.
\item We show via extensive experiments that our proposed approach significantly improves the effectiveness of semantic segmentation models while achieving competitive performance on classical road defect detection tasks.
\end{itemize}

\section{Related Work}
\subsection{Semantic segmentation}
In the domain of computer vision, semantic segmentation, the task of assigning a class label to each pixel in an image to facilitate comprehensive scene understanding, has emerged as a crucial research area. It has garnered significant attention within the academic community, leading to the development of a plethora of methodologies aimed at enhancing its accuracy and efficiency.

Early approaches to semantic segmentation used handcrafted features and classification algorithms to assign labels to image pixels. For example, the seminal work of Coleman~\cite{coleman1979image} used clustering to segment images into object regions, and is based on a mathematical-pattern recognition model. 
In addition, there are a series of works~\cite{kaur2014various,garcia2018segmentation,horvath2006image} using semantic segmentation based on basic features. However, these approaches were limited by their reliance on handcrafted features, which were not robust to changes in image content or lighting conditions.
A prominent approach involves leveraging deep learning models, specifically convolutional neural networks (CNNs)~\cite{schmidhuber2015deep}, which have demonstrated remarkable success in extracting complex hierarchical features from images. Further advancements have been made through the adoption of fully convolutional networks (FCNs)~\cite{long2015fully}, which enable end-to-end training and efficient inference for pixel-wise classification.
More recently, deep learning techniques, especially convolutional neural networks (CNNs), have become the dominant approach for semantic segmentation. Long {\it et al.}~\cite{long2015fully} proposed fully convolutional networks (FCNs) that can process images of arbitrary sizes and output pixel-level segmentation maps.    Since then, many variations and extensions of FCNs have been proposed, including U-Net \cite{ronneberger2015u}, SegNet \cite{badrinarayanan2017segnet}, and DeepLab \cite{chen2014semantic}.

As the field continues to evolve, researchers are investigating novel techniques such as incorporating domain adaptation\cite{li2019bidirectional}, attention mechanisms\cite{azad2020attention}, and generative adversarial networks (GANs) \cite{souly2017semi} to refine the performance of semantic segmentation models. The unrelenting pursuit of innovation in this area has led to substantial improvements in a wide range of applications, including autonomous navigation, medical imaging, and remote sensing, underscoring the importance of semantic segmentation in the broader landscape of artificial intelligence.
In addition to the development of new network architectures, researchers have also explored various techniques to improve the accuracy and efficiency of semantic segmentation. For example, Chen et al.~\cite{chen2017rethinking} introduced atrous convolution, which enables the use of large receptive fields without increasing the number of network parameters. Similarly, Yu et al.~\cite{yu2015multi} proposed the use of dilated convolution, which allows for multi-scale processing of images.
Another area of research has been the incorporation of contextual information into the segmentation process. Various techniques have been proposed to capture global context, including spatial pyramid pooling \cite{he2015spatial}. Wang et al. \cite{wang2018dense} introduced the use of the atrous spatial pyramid pooling (ASPP) module, which uses atrous convolution at multiple scales to capture multi-scale context.
Finally, there has been increasing interest in the use of weakly supervised or unsupervised approaches to semantic segmentation, which can alleviate the need for pixel-level annotations. For example, Huang et al. \cite{huang2018weakly} proposed a weakly supervised approach that uses image-level labels to train a CNN for semantic segmentation. Similarly, Kalluri et al. \cite{kalluri2019universal} proposed an unsupervised approach that uses a deep generative model to learn a representation that can be used for semantic segmentation.

\subsection{Adversarial learning}
Adversarial learning is an active area of research in machine learning that aims to improve the robustness and generalization of models by explicitly training them to defend against adversarial attacks. The study of adversarial learning is predicated on the dynamic interplay between two competing entities: the learner, whose objective is to generate accurate predictions, and the adversary, whose goal is to undermine those predictions through the introduction of subtle input perturbations. Groundbreaking research in this domain has given rise to a multitude of defense strategies, such as adversarial training, gradient obfuscation, and input preprocessing techniques.

One of the earliest approaches to adversarial learning was proposed by Goodfellow et al. \cite{goodfellow2020generative}, who introduced the concept of adversarial examples – inputs that are intentionally perturbed to cause misclassification by a neural network. Since then, a variety of techniques have been developed to generate adversarial examples, including Fast Gradient Sign Method (FGSM) \cite{liu2019sensitivity}, and Iterative FGSM \cite{dong2018boosting}.
To defend against adversarial attacks, researchers have proposed various techniques based on adversarial training. In adversarial training, a model is trained on a mixture of clean and adversarial examples, with the goal of making the model more robust to attacks. Balaji et al. \cite{balaji2020robust} introduced a robust optimization approach, which maximizes the worst-case loss over a set of perturbations within a given distance from the input. Similarly, Bao et al. \cite{bao2020coupling} proposed the use of random smoothing, which adds noise to the input to make it more robust to adversarial perturbations.
Another approach to adversarial learning is to use generative models to learn the distribution of the data and generate realistic samples. This can be useful in scenarios where the training data is limited or difficult to obtain. For example, Generative Adversarial Networks have been used to generate adversarial examples and to augment training data for improving the robustness of models.
Recent work has also explored the use of adversarial learning in other areas of machine learning, such as reinforcement learning and transfer learning. For example, Zhao et al. \cite{zhao2021model} proposed adversarial imitation learning, which uses adversarial training to learn policies that are robust to adversarial perturbations. Similarly, Soleimani et al. \cite{soleimani2021cross} proposed adversarial transfer learning, which uses adversarial examples to transfer knowledge between related tasks.

\section{Methodology}
The semantic segmentation network framework, exemplified by Unet, primarily encompasses the encoder-decoder feature pipeline and inter-level connectivity channels. The encoder-decoder framework involves the downsampling and upsampling of input feature maps, with the aim of generating semantic information at multiple scales. Nevertheless, excessive downsampling and upsampling operations may result in the loss of partial information, hindering the decoder's ability to produce desirable outputs. Consequently, the utilization of adversarial learning techniques to enhance the generator's performance has gained widespread practical application.

To address this, we propose a network framework that combines both architectures, employing adversarial learning to augment the decoder's performance. In this approach, we incorporate a discriminator from adversarial learning for each layer of the decoder, supervising the decoder's performance and training the encoder levels accordingly.
\begin{figure}[!htbp]
\centering 
\includegraphics[width = 0.45\textwidth]{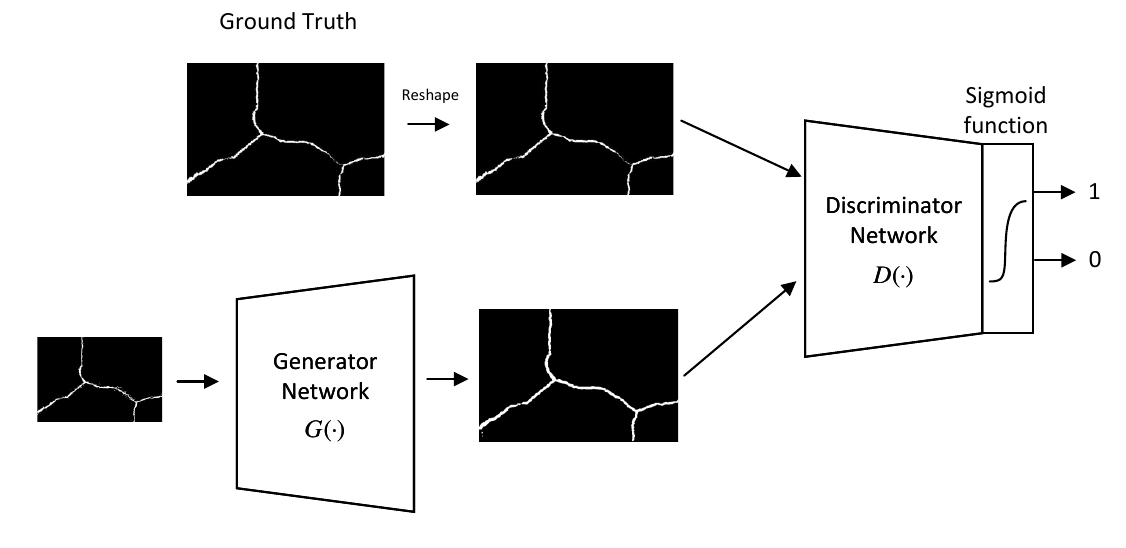}
\caption{To incorporate adversarial learning into the generator, we introduce a step where the output of each generator is discriminated against the ground truth. The discriminator takes two inputs: the ground truth annotations of the faulty images and the output of the generator. The generator, in turn, performs upsampling on the input images, transforming low-resolution images into high-resolution ones. The discriminator's task is to discern the similarity between these two inputs.}
\label{DG}
\end{figure}

\begin{figure*}[!htbp]
\centering 
\includegraphics[width = 0.9\textwidth]{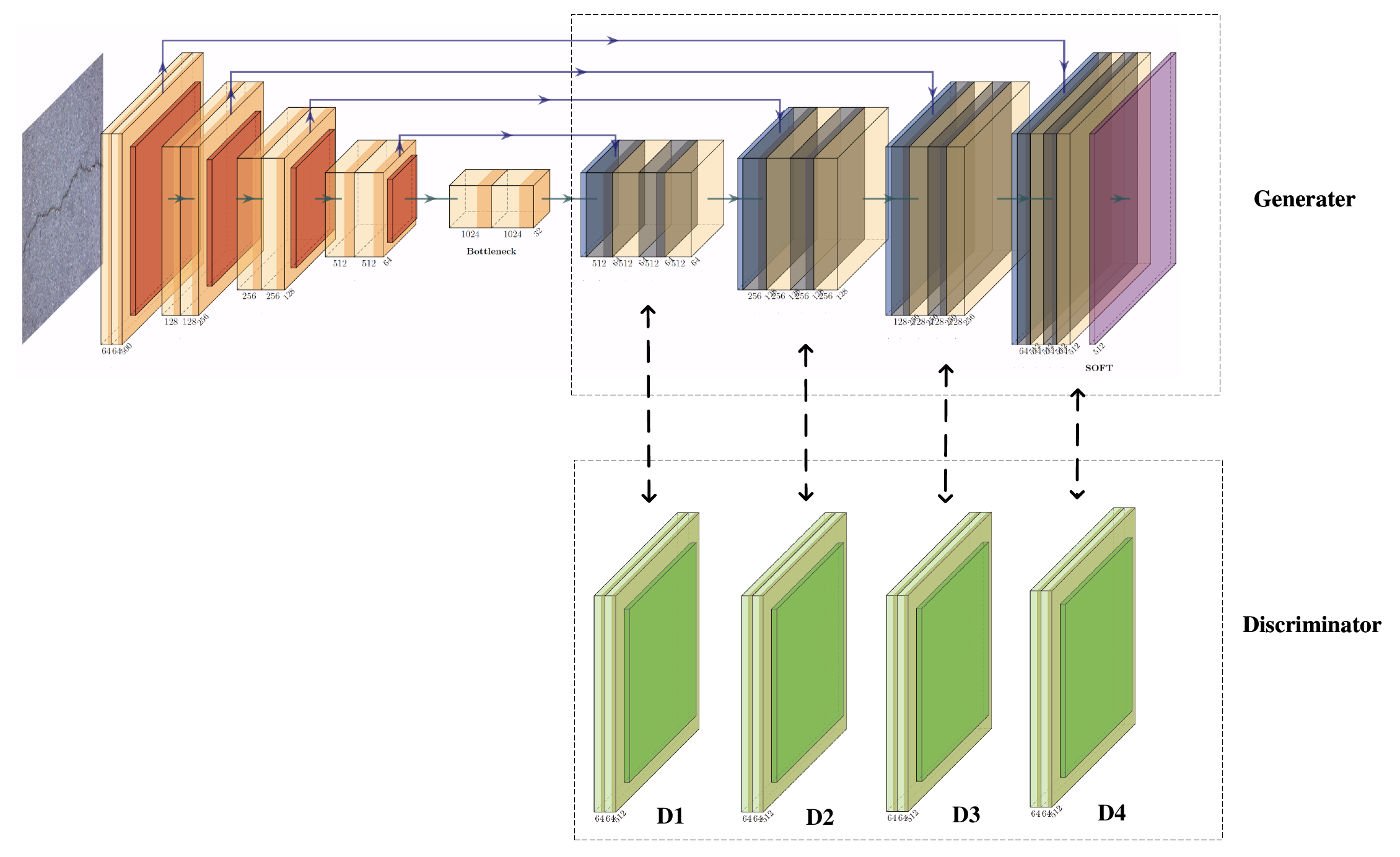}
\caption{Illustration of our approach: 
The entire upper section represents an Unet network structure, wherein we have incorporated a discriminator at each generator within the Unet. Each discriminator follows the same structure and is responsible for distinguishing the output images at that particular layer. Multiple discriminators work in parallel to form a collective discriminator.}
\label{fig_overall}
\end{figure*}

\subsection{Problem Formulation} 

The problem formulation of semantic segmentation can be summarized as follows: Given an input image, develop a model that can accurately assign a class label to each pixel, allowing for the understanding and interpretation of the scene at a granular level. The primary objective of semantic segmentation is to assign a class label to each pixel in an input image, such that each pixel belongs to a specific object or region in the image. This process allows for the understanding and analysis of the image at a more granular level, enabling applications such as autonomous driving, robotics, medical image analysis, and more.

Input: The input to the semantic segmentation problem is typically a digital image represented as a matrix of pixel values, with dimensions $height ($H$) \times width ($W$) \times channels ($C$)$, where channels could be RGB ($3$ channels) for color images or grayscale ($1$ channel) for black and white images.

Output: The output is a segmented image of the same dimensions as the input image $($H$ \times $W$)$,where each pixel is assigned a class label representing the object or region it belongs to. The number of classes depends on the specific problem and dataset used.

\subsection{Codec structure framework with Adversarial learning}

The encoder structure in the codec architecture is actually a kind of inverse convolution operation, which is divided into three steps: 1) The input feature map undergoes certain transformations to yield a new feature map; 2) Obtain a new convolution kernel setting to obtain a new convolution kernel setting; 3) The new convolutional filters are convolved with the transformed feature map in a conventional manner, resulting in the output of the deconvolution operation and yielding the desired outcome.

During this process, we incorporate a step of adversarial learning, where the output of each generator is discriminated against the ground truth, as illustrated in Fig.~\ref{DG}. 
The discriminator in this setup takes two inputs: the ground truth annotations of the images depicting faults and the output of the generator. The generator is responsible for upsampling the input images, transforming low-resolution images into high-resolution ones. The discriminator's role is to determine the similarity between the two inputs. It consists of a combination of CNN (Convolutional Neural Network) and BN (Batch Normalization) layers. The final layer of the discriminator is equipped with a Sigmoid function to ensure regularization, producing a similarity score between $0$ and $1$.

When dealing with different codec architectures, some may feature a cascaded structure with multiple generators. In such cases, a structure comprising multiple generator-discriminator pairs is required. Taking Unet as an example, we demonstrate the incorporation of discriminators within a multi-layer generator network, as shown in Fig.~\ref{fig_overall}. The entire upper section represents an Unet network structure, with a discriminator added at each generator within the Unet. Each discriminator follows an identical structure, responsible for discriminating the output images at its corresponding layer. Multiple discriminators work in parallel, forming an overall discriminator.

Each discriminator outputs a mean error, and when n discriminators operate in parallel, we obtain n mean errors and n Dice Loss. These mean errors are then added to the original generator's loss to derive the loss for the new network structure. 
Given the n generators and discriminators, the task-tailored loss function can be defined as:
\begin{equation}
\label{equ:1}
{{\cal L}_{new}} = {\cal L}_{ori} + \sum\limits_{i = 1}^n (MES_{n}+DiceLoss_{n}).
\end{equation} 

It is important to note that the ground truth input for each discriminator at different layers is not the same. Only the last layer's discriminator takes the ground truth original image as input. In contrast, the input for the discriminators at previous layers consists of downsampled versions of the ground truth image. The purpose of downsampling is to ensure that the size of the downsampled image matches the output of the corresponding generator in that layer.

\section{Experimental Results}
\subsection{Dataset and Evaluation Protocol.}
\textbf{Dataset.}In order to evaluate the efficacy of our proposed methodology within the domain of Intelligent Transportation Systems (ITS), we conducted a comparative analysis against established semantic segmentation techniques widely employed in ITS datasets. Specifically, we benchmarked our approach against two prominent methods, namely CrackForest \cite{shi2016automatic,cui2015pavement} and Deepcrack \cite{liu2019deepcrack}.

\textbf{Evaluation Metrics}
To assess the quality of the segmentation results, evaluation metrics are employed. Commonly used evaluation metrics for semantic segmentation include Mean Intersection over Union (mIoU) and Dice coefficient (Dice). The Mean Intersection over Union (mIoU) is a performance measure that evaluates the overlap between the predicted segmentation mask and the ground truth mask across multiple classes. It calculates the average of the Intersection over Union (IoU) scores for each class, providing an overall assessment of the model's segmentation accuracy. A higher mIoU value indicates better segmentation performance.
The Dice coefficient, also known as the Sørensen-Dice coefficient, is another widely used metric for evaluating the similarity between the predicted and ground truth segmentation masks. It quantifies the overlap between the two masks by computing the ratio of the intersection to the sum of the areas of the predicted and ground truth regions. The Dice coefficient ranges from $0$ to $1$, with $1$ indicating a perfect match with the labels.
These evaluation metrics, mIoU and Dice, serve as objective measures to assess the quality and accuracy of the semantic segmentation models in quantifying the degree of agreement between the predicted and ground truth masks.

\subsection{Implementation Details}
The experimentation phase of this research study was carried out on the CentOS 7 operating system. Our proposed method was trained and evaluated within a Python 3.8 environment, employing the PyTorch software package. The computations were accelerated using the Nvidia GeForce RTX A5000 GPU.

\subsection{Performance Study}
In our experimental section, we compared three baseline models, namely SEGAN~\cite{zhang2018seggan}, Unet, and Nest, which are commonly used in semantic segmentation algorithms. SEGAN combines generative adversarial networks (GANs) with the capabilities of denoising autoencoders to enhance the quality and intelligibility of noisy speech signals.
Unet is a widely used neural network architecture for image segmentation tasks. It follows an encoder-decoder structure, where the encoder extracts high-level features from the input image, and the decoder performs upsampling to generate a segmentation map.
Nest, also known as Nested Unet, is an extension of the Unet architecture that incorporates skip connections between multiple levels of the encoder and decoder. This nested structure allows for the integration of more fine-grained details and contextual information during the upsampling process, resulting in improved segmentation accuracy.

\noindent\textbf{Quantitative Comparison}
We utilize a supervision strategy for low resolution outputs to enhance the high resolution output effect. Specifically, the combined ratio of loss functions is a crucial parameter set that significantly impacts the model performance. Therefore, we conducted an experimental analysis of the combined ratio of loss functions, and the results are presented in a table. The findings indicate that high attention coefficients should be given to high-resolution outputs and vice versa. In a specific combination ratio, any increase in the supervision for low resolution outputs led to a negative impact on model performance, as shown in rows 3 and 4 of the table. Similarly, any decrease in supervision on high-resolution output resulted in difficulty achieving optimal model performance, as seen in rows 4 and 5 of the table. Ultimately, we determined that the optimal model performance is achieved when l1=1, l2=0.3, l3=0.1, l4=0.05, and l5=0.01.

\noindent\textbf{Ablation Study}
We perform an in-depth study of the proposed framework, focusing on the degree of deep supervision of the generators in the generator framework, and conduct ablation experiments. Specifically, the effect on segmentation performance with different resolution outputs of the decoder in the generator is investigated, as shown in Table. The experimental results show that the introduction of the low-resolution output gives better performance to the model (mIoU improves from 0.5808 to 0.6422). This result proves that as the degree of deep supervision increases, the decoder of the generator introduces more information and constraints that are useful for the performance of the model
\begin{table}[!t]
\caption{
Experimental Results on CrackForest Dataset
}
\label{tab:CrackForest}
\centering
\small
\begin{tabular}{lccc} 
\hline
{Methods}   &   &  {mIoU} &  {Dice}    \\
\hline
SEGAN   &   &  0.5808   & 0.6746 \\
Unet   &   &  0.6053   & 0.7154 \\
Nest   &   & 0.6305  & 0.7459 \\
\hline
Ours     &   &  \textbf{0.6422}   & \textbf{0.7539} \\
\hline
\end{tabular}
\end{table}

\begin{table}[!t]
\caption{
Experimental Results on Deepcrast Dataset
}
\label{tab:Deepcrast}
\centering
\small
\begin{tabular}{lccc} 
\hline
{Methods}   &   &  {mIoU} &  {Dice}    \\
\hline
SEGAN   &   &  0.4574   & 0.6120 \\
Unet   &   &  0.5022   & 0.6536 \\
Nest   &   & 0.5127  & 0.6620 \\
\hline
Ours     &   &  \textbf{0.5172}   & \textbf{0.6735} \\
\hline
\end{tabular}
\end{table}

\begin{table}[!t]
\caption{
Ablation study
}
\label{tab:Ablation}
\centering
\small
\resizebox{\linewidth}{!}{
\begin{tabular}{lccccccc} 
\hline
\multicolumn{5}{c}{Combination of loss coefficients}                        &   {mIoU} &  {Dice}    \\
\hline
l1=1   &  l2=0   &  l3=0   & l4=0     & l5=0     &  0.5808   & 0.6746 \\
l1=1   &  l2=0.3 &  l3=0   & l4=0     & l5=0     &  0.5910   & 0.6861 \\
l1=1   &  l2=0.3 &  l3=0.1 & l4=0     & l5=0     &  0.5795   & 0.6710 \\
l1=1   &  l2=0.3 &  l3=0.1 & l3=0.05  & l5=0     &  0.5977   & 0.6851 \\
l1=1   &  l2=0.3 &  l3=0.1 & l3=0.05  & l5=0.01  &  0.6422   & 0.7539 \\

\hline
\end{tabular}}
\end{table}

\begin{table}[!t]
\caption{
Parameter experiment
}
\label{tab:Parameter experiment}
\centering
\small
\resizebox{\linewidth}{!}{
\begin{tabular}{lccccccc} 
\hline
\multicolumn{5}{c}{Combination of loss coefficients}        &  {mIoU} &  {Dice}    \\
\hline
l1=1   &  l2=0.8 &  l3=0.6 & l4=0.4   & l5=0.2   &  0.5913   & 0.6811 \\
l1=1   &  l2=0.5 &  l3=0.2 & l4=0.1   & l5=0.05  &  0.6087   & 0.7023 \\
l1=1   &  l2=0.3 &  l3=0.25& l3=0.15  & l5=0.1   &  0.5947   & 0.7002 \\
l1=1   &  l2=0.3 &  l3=0.1 & l4=0.05  & l5=0.01  &  0.6422   & 0.7539 \\
l1=1   &  l2=0.2 &  l3=0.1 & l4=0.05  & l5=0.01  &  0.6066   & 0.6989 \\

\hline
\end{tabular}}
\end{table}

\section{Conclusions}
In this paper, we propose a network structure that combines a codec structure framework and adversarial learning, which improves the effectiveness of the generators in the original network with several additional adversarial learning networks. Experimental results show that the method is effective in improving the generator capability of the codec network and generating more accurate semantic segmentation results. Quantitative evaluation results on several standard datasets and ablation studies validate that our proposed method outperforms many existing state-of-the-art methods and shows great effectiveness and potential in ITS.
\bibliographystyle{IEEEtran}
\bibliography{ref}

\begin{thebibliography}{10}
\providecommand{\url}[1]{#1}
\csname url@samestyle\endcsname
\providecommand{\newblock}{\relax}
\providecommand{\bibinfo}[2]{#2}
\providecommand{\BIBentrySTDinterwordspacing}{\spaceskip=0pt\relax}
\providecommand{\BIBentryALTinterwordstretchfactor}{4}
\providecommand{\BIBentryALTinterwordspacing}{\spaceskip=\fontdimen2\font plus
\BIBentryALTinterwordstretchfactor\fontdimen3\font minus
  \fontdimen4\font\relax}
\providecommand{\BIBforeignlanguage}[2]{{%
\expandafter\ifx\csname l@#1\endcsname\relax
\typeout{** WARNING: IEEEtran.bst: No hyphenation pattern has been}%
\typeout{** loaded for the language `#1'. Using the pattern for}%
\typeout{** the default language instead.}%
\else
\language=\csname l@#1\endcsname
\fi
#2}}
\providecommand{\BIBdecl}{\relax}
\BIBdecl

\bibitem{figueiredo2001towards}
L.~Figueiredo, I.~Jesus, J.~T. Machado, J.~R. Ferreira, and J.~M. De~Carvalho,
  ``Towards the development of intelligent transportation systems,'' \emph{ITSC
  2001. 2001 IEEE intelligent transportation systems. Proceedings (Cat. No.
  01TH8585)}, pp. 1206--1211, 2001.

\bibitem{papadimitratos2009vehicular}
P.~Papadimitratos, A.~De~La~Fortelle, K.~Evenssen, R.~Brignolo, and S.~Cosenza,
  ``Vehicular communication systems: Enabling technologies, applications, and
  future outlook on intelligent transportation,'' \emph{IEEE communications
  magazine}, vol.~47, no.~11, pp. 84--95, 2009.

\bibitem{cc1}
C.~Chen, K.~Li, S.~G. Teo, X.~Zou, K.~Li, and Z.~Zeng, ``Citywide traffic flow
  prediction based on multiple gated spatio-temporal convolutional neural
  networks,'' \emph{ACM Transactions on Knowledge Discovery from Data (TKDD)},
  vol.~14, no.~4, pp. 1--23, 2020.

\bibitem{oztel2021vision}
G.~Y. Oztel, ``Vision-based road segmentation for intelligent vehicles using
  deep convolutional neural networks,'' in \emph{2021 International Conference
  on INnovations in Intelligent SysTems and Applications (INISTA)}.\hskip 1em
  plus 0.5em minus 0.4em\relax IEEE, 2021, pp. 1--5.

\bibitem{wan2020intelligent}
S.~Wan, X.~Xu, T.~Wang, and Z.~Gu, ``An intelligent video analysis method for
  abnormal event detection in intelligent transportation systems,'' \emph{IEEE
  Transactions on Intelligent Transportation Systems}, vol.~22, no.~7, pp.
  4487--4495, 2020.

\bibitem{kaushik2020leveraging}
S.~Kaushik, A.~Raman, and K.~R. Rao, ``Leveraging computer vision for emergency
  vehicle detection-implementation and analysis,'' in \emph{2020 11th
  International Conference on Computing, Communication and Networking
  Technologies (ICCCNT)}.\hskip 1em plus 0.5em minus 0.4em\relax IEEE, 2020,
  pp. 1--6.

\bibitem{othman2019road}
Z.~Othman, A.~Abdullah, F.~Kasmin, and S.~S.~S. Ahmad, ``Road crack detection
  using adaptive multi resolution thresholding techniques,'' \emph{TELKOMNIKA
  (Telecommunication Computing Electronics and Control)}, vol.~17, no.~4, pp.
  1874--1881, 2019.

\bibitem{singh2021highway}
R.~Singh, R.~Sharma, S.~V. Akram, A.~Gehlot, D.~Buddhi, P.~K. Malik, and
  R.~Arya, ``Highway 4.0: Digitalization of highways for vulnerable road safety
  development with intelligent iot sensors and machine learning,'' \emph{Safety
  science}, vol. 143, p. 105407, 2021.

\bibitem{arena2020review}
F.~Arena, G.~Pau, and A.~Severino, ``A review on ieee 802.11 p for intelligent
  transportation systems,'' \emph{Journal of Sensor and Actuator Networks},
  vol.~9, no.~2, p.~22, 2020.

\bibitem{qi2017pointnet}
C.~R. Qi, H.~Su, K.~Mo, and L.~J. Guibas, ``Pointnet: Deep learning on point
  sets for 3d classification and segmentation,'' in \emph{Proceedings of the
  IEEE conference on computer vision and pattern recognition}, 2017, pp.
  652--660.

\bibitem{li2018pointcnn}
Y.~Li, R.~Bu, M.~Sun, W.~Wu, X.~Di, and B.~Chen, ``Pointcnn: Convolution on
  x-transformed points,'' \emph{Advances in neural information processing
  systems}, vol.~31, 2018.

\bibitem{cc2}
C.~Chen, K.~Li, W.~Wei, J.~T. Zhou, and Z.~Zeng, ``Hierarchical graph neural
  networks for few-shot learning,'' \emph{IEEE Transactions on Circuits and
  Systems for Video Technology}, vol.~32, no.~1, pp. 240--252, 2021.

\bibitem{kashyap2022traffic}
A.~A. Kashyap, S.~Raviraj, A.~Devarakonda, S.~R. Nayak~K, S.~KV, and S.~J.
  Bhat, ``Traffic flow prediction models--a review of deep learning
  techniques,'' \emph{Cogent Engineering}, vol.~9, no.~1, p. 2010510, 2022.

\bibitem{coleman1979image}
G.~B. Coleman and H.~C. Andrews, ``Image segmentation by clustering,''
  \emph{Proceedings of the IEEE}, vol.~67, no.~5, pp. 773--785, 1979.

\bibitem{kaur2014various}
D.~Kaur and Y.~Kaur, ``Various image segmentation techniques: a review,''
  \emph{International Journal of Computer Science and Mobile Computing},
  vol.~3, no.~5, pp. 809--814, 2014.

\bibitem{garcia2018segmentation}
F.~Garcia-Lamont, J.~Cervantes, A.~L{\'o}pez, and L.~Rodriguez, ``Segmentation
  of images by color features: A survey,'' \emph{Neurocomputing}, vol. 292, pp.
  1--27, 2018.

\bibitem{horvath2006image}
J.~Horvath, ``Image segmentation using fuzzy c-means,'' in \emph{Symposium on
  Applied Machine Intelligence}, 2006.

\bibitem{schmidhuber2015deep}
J.~Schmidhuber, ``Deep learning in neural networks: An overview,'' \emph{Neural
  networks}, vol.~61, pp. 85--117, 2015.

\bibitem{long2015fully}
J.~Long, E.~Shelhamer, and T.~Darrell, ``Fully convolutional networks for
  semantic segmentation,'' in \emph{Proceedings of the IEEE conference on
  computer vision and pattern recognition}, 2015, pp. 3431--3440.

\bibitem{ronneberger2015u}
O.~Ronneberger, P.~Fischer, and T.~Brox, ``U-net: Convolutional networks for
  biomedical image segmentation,'' in \emph{Medical Image Computing and
  Computer-Assisted Intervention--MICCAI 2015: 18th International Conference,
  Munich, Germany, October 5-9, 2015, Proceedings, Part III 18}.\hskip 1em plus
  0.5em minus 0.4em\relax Springer, 2015, pp. 234--241.

\bibitem{badrinarayanan2017segnet}
V.~Badrinarayanan, A.~Kendall, and R.~Cipolla, ``Segnet: A deep convolutional
  encoder-decoder architecture for image segmentation,'' \emph{IEEE
  transactions on pattern analysis and machine intelligence}, vol.~39, no.~12,
  pp. 2481--2495, 2017.

\bibitem{chen2014semantic}
L.-C. Chen, G.~Papandreou, I.~Kokkinos, K.~Murphy, and A.~L. Yuille, ``Semantic
  image segmentation with deep convolutional nets and fully connected crfs,''
  \emph{arXiv preprint arXiv:1412.7062}, 2014.

\bibitem{li2019bidirectional}
Y.~Li, L.~Yuan, and N.~Vasconcelos, ``Bidirectional learning for domain
  adaptation of semantic segmentation,'' in \emph{Proceedings of the IEEE/CVF
  Conference on Computer Vision and Pattern Recognition}, 2019, pp. 6936--6945.

\bibitem{azad2020attention}
R.~Azad, M.~Asadi-Aghbolaghi, M.~Fathy, and S.~Escalera, ``Attention
  deeplabv3+: Multi-level context attention mechanism for skin lesion
  segmentation,'' in \emph{Computer Vision--ECCV 2020 Workshops: Glasgow, UK,
  August 23--28, 2020, Proceedings, Part I 16}.\hskip 1em plus 0.5em minus
  0.4em\relax Springer, 2020, pp. 251--266.

\bibitem{souly2017semi}
N.~Souly, C.~Spampinato, and M.~Shah, ``Semi supervised semantic segmentation
  using generative adversarial network,'' in \emph{Proceedings of the IEEE
  international conference on computer vision}, 2017, pp. 5688--5696.

\bibitem{chen2017rethinking}
L.-C. Chen, G.~Papandreou, F.~Schroff, and H.~Adam, ``Rethinking atrous
  convolution for semantic image segmentation,'' \emph{arXiv preprint
  arXiv:1706.05587}, 2017.

\bibitem{yu2015multi}
F.~Yu and V.~Koltun, ``Multi-scale context aggregation by dilated
  convolutions,'' \emph{arXiv preprint arXiv:1511.07122}, 2015.

\bibitem{he2015spatial}
K.~He, X.~Zhang, S.~Ren, and J.~Sun, ``Spatial pyramid pooling in deep
  convolutional networks for visual recognition,'' \emph{IEEE transactions on
  pattern analysis and machine intelligence}, vol.~37, no.~9, pp. 1904--1916,
  2015.

\bibitem{wang2018dense}
Y.~Wang, B.~Liang, M.~Ding, and J.~Li, ``Dense semantic labeling with atrous
  spatial pyramid pooling and decoder for high-resolution remote sensing
  imagery,'' \emph{Remote Sensing}, vol.~11, no.~1, p.~20, 2018.

\bibitem{huang2018weakly}
Z.~Huang, X.~Wang, J.~Wang, W.~Liu, and J.~Wang, ``Weakly-supervised semantic
  segmentation network with deep seeded region growing,'' in \emph{Proceedings
  of the IEEE conference on computer vision and pattern recognition}, 2018, pp.
  7014--7023.

\bibitem{kalluri2019universal}
T.~Kalluri, G.~Varma, M.~Chandraker, and C.~Jawahar, ``Universal
  semi-supervised semantic segmentation,'' in \emph{Proceedings of the IEEE/CVF
  International Conference on Computer Vision}, 2019, pp. 5259--5270.

\bibitem{goodfellow2020generative}
I.~Goodfellow, J.~Pouget-Abadie, M.~Mirza, B.~Xu, D.~Warde-Farley, S.~Ozair,
  A.~Courville, and Y.~Bengio, ``Generative adversarial networks,''
  \emph{Communications of the ACM}, vol.~63, no.~11, pp. 139--144, 2020.

\bibitem{liu2019sensitivity}
Y.~Liu, S.~Mao, X.~Mei, T.~Yang, and X.~Zhao, ``Sensitivity of adversarial
  perturbation in fast gradient sign method,'' in \emph{2019 IEEE symposium
  series on computational intelligence (SSCI)}.\hskip 1em plus 0.5em minus
  0.4em\relax IEEE, 2019, pp. 433--436.

\bibitem{dong2018boosting}
Y.~Dong, F.~Liao, T.~Pang, H.~Su, J.~Zhu, X.~Hu, and J.~Li, ``Boosting
  adversarial attacks with momentum,'' in \emph{Proceedings of the IEEE
  conference on computer vision and pattern recognition}, 2018, pp. 9185--9193.

\bibitem{balaji2020robust}
Y.~Balaji, R.~Chellappa, and S.~Feizi, ``Robust optimal transport with
  applications in generative modeling and domain adaptation,'' \emph{Advances
  in Neural Information Processing Systems}, vol.~33, pp. 12\,934--12\,944,
  2020.

\bibitem{bao2020coupling}
J.~Bao, L.~Li, and F.~Redoloza, ``Coupling ensemble smoother and deep learning
  with generative adversarial networks to deal with non-gaussianity in flow and
  transport data assimilation,'' \emph{Journal of Hydrology}, vol. 590, p.
  125443, 2020.

\bibitem{zhao2021model}
T.~Zhao, Y.~Wang, G.~Li, L.~Kong, Y.~Chen, Y.~Wang, N.~Xie, and J.~Yang, ``A
  model-based reinforcement learning method based on conditional generative
  adversarial networks,'' \emph{Pattern Recognition Letters}, vol. 152, pp.
  18--25, 2021.

\bibitem{soleimani2021cross}
E.~Soleimani and E.~Nazerfard, ``Cross-subject transfer learning in human
  activity recognition systems using generative adversarial networks,''
  \emph{Neurocomputing}, vol. 426, pp. 26--34, 2021.

\bibitem{shi2016automatic}
Y.~Shi, L.~Cui, Z.~Qi, F.~Meng, and Z.~Chen, ``Automatic road crack detection
  using random structured forests,'' \emph{IEEE Transactions on Intelligent
  Transportation Systems}, vol.~17, no.~12, pp. 3434--3445, 2016.

\bibitem{cui2015pavement}
L.~Cui, Z.~Qi, Z.~Chen, F.~Meng, and Y.~Shi, ``Pavement distress detection
  using random decision forests,'' in \emph{International Conference on Data
  Science}.\hskip 1em plus 0.5em minus 0.4em\relax Springer, 2015, pp. 95--102.

\bibitem{liu2019deepcrack}
Y.~Liu, J.~Yao, X.~Lu, R.~Xie, and L.~Li, ``Deepcrack: A deep hierarchical
  feature learning architecture for crack segmentation,''
  \emph{Neurocomputing}, vol. 338, pp. 139--153, 2019.

\bibitem{zhang2018seggan}
X.~Zhang, X.~Zhu, N.~Zhang, P.~Li, L.~Wang \emph{et~al.}, ``Seggan: Semantic
  segmentation with generative adversarial network,'' in \emph{2018 IEEE Fourth
  International Conference on Multimedia Big Data (BigMM)}.\hskip 1em plus
  0.5em minus 0.4em\relax IEEE, 2018, pp. 1--5.

\end{thebibliography}

\end{document}